# Patterns

# Improving molecular representation learning with metric learning-enhanced optimal transport

## Graphical abstract

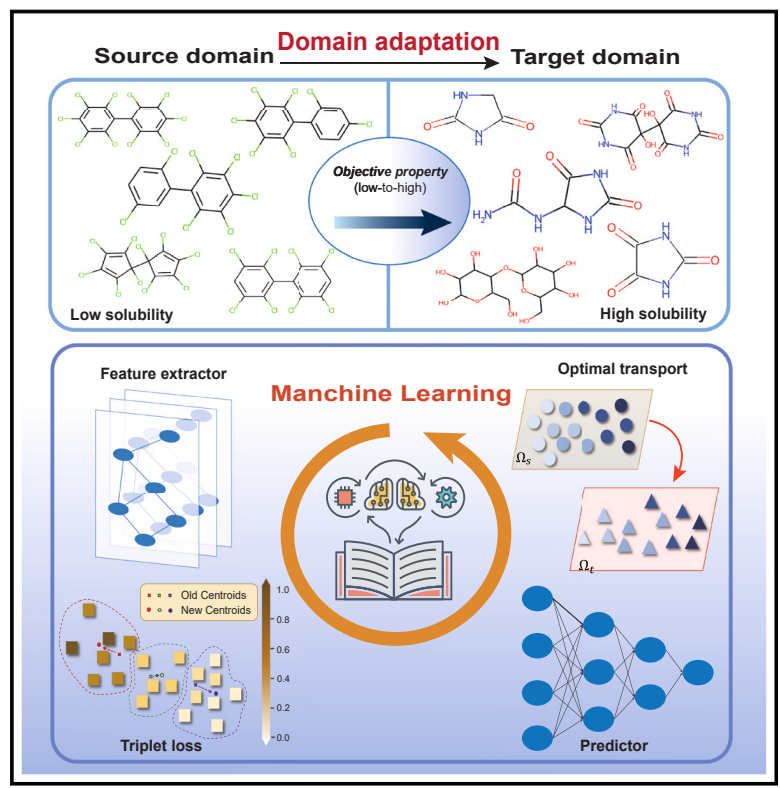

## Highlights

- We bring up the domain adaptation problem in molecular representation learning

- We design an optimal transport method for chemical domain adaptation

- MROT outperforms all baselines in predictive tasks of molecules and materials

- This approach may help improve the generality of other deep-learning models

## Authors

Fang Wu, Nicolas Courty, Shuting Jin, Stan Z. Li

## Correspondence
stan.zq.li@westlake.edu.cn

## In brief

Molecular representation learning has recently emerged as an essential tool for fields such as drug discovery and material synthesis. However, it is still a big challenge to overcome the limitation and heterogeneity of training data for more efficient molecular representation learning. To this end, we propose an optimal transport strategy dubbed MORT. MORT can achieve excellent performance in many molecular regression tasks of unsupervised and semi-supervised domain adaptation and can solve prediction problems beyond the range of training samples.

# Patterns

## Article

# Improving molecular representation learning with metric learning-enhanced optimal transport

Fang Wu,[1,4,6] Nicolas Courty,[2,6] Shuting Jin,[3,5,6] and Stan Z. Li[1,7,*]

[1]School of Engineering, Westlake University, Hangzhou 310024, China
[2]French National Centre for Scientific Research, Southern Brittany University, Lorient, France
[3]School of Informatics, Xiamen University, Xiamen 361005, China
[4]Institute of AI Industry Research, Tsinghua University, Beijing 100084, China
[5]National Institute for Data Science in Health and Medicine, Xiamen University, Xiamen 361005, China
[6]These authors contributed equally
[7]Lead contact
*Correspondence: stan.zq.li@westlake.edu.cn
https://doi.org/10.1016/j.patter.2023.100714

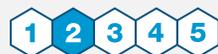

**THE BIGGER PICTURE** Molecular representation learning is significant for drug discovery and material synthesis. Training data in computational science are usually limited, requiring predictive models to have predictive capabilities beyond the scope of training samples. Domain adaptation (DA) is a field associated with machine learning that aligns the data distributions of the source (training dataset) and target (test dataset) domains by learning a more generalized feature space. In this article, we propose an optimal transport-based algorithm named MROT to overcome this DA challenge and improve the generalization capacity of predictive models. Our study shows that our MROT framework provides a potential solution for basic chemical regression prediction tasks, including chemical property prediction and materials adsorption selection. This work contributes a new idea to achieve the proper transformation from simulated data to experimental data, which play a crucial role in exploring new drugs or materials.

**Proof-of-Concept:** Data science output has been formulated, implemented, and tested for one domain/problem

## SUMMARY

Training data are usually limited or heterogeneous in many chemical and biological applications. Existing machine learning models for chemistry and materials science fail to consider generalizing beyond training domains. In this article, we develop a novel optimal transport-based algorithm termed MROT to enhance their generalization capability for molecular regression problems. MROT learns a continuous label of the data by measuring a new metric of domain distances and a posterior variance regularization over the transport plan to bridge the chemical domain gap. Among downstream tasks, we consider basic chemical regression tasks in unsupervised and semi-supervised settings, including chemical property prediction and materials adsorption selection. Extensive experiments show that MROT significantly outperforms state-of-the-art models, showing promising potential in accelerating the discovery of new substances with desired properties.

## INTRODUCTION

It has become a growing trend to leverage deep learning (DL) in computational chemistry.[1] Traditionally, DL techniques assume that the training data (e.g., simulated data) and the testing data (e.g., experimental data) come from the same distribution, so models are expected to perform well on the testing data.[2–4] Nevertheless, real-world data for learning and inference are unlikely to follow the same distribution, and models usually fail to perform extraordinarily when generalizing to data of unseen domains. Especially in many chemical applications, training data are necessarily limited or otherwise heterogeneous to the testing data. For instance, learning from one category of molecules and deploying an application targeted to a wide range of other groups may be hindered by different distributions of their 3D constructions and atom compositions.[5] For the reasons mentioned above, a trustworthy chemical DL system should not only produce accurate predictions on the known compounds but possess the ability to transfer knowledge across domains.[6] This adaptation endows chemists with the power to find new potential materials and drugs,[7] and the scientific research process can be substantially accelerated.

Domain adaptation (DA) is a subfield in machine learning that aims to cope with this type of problem by bridging the gaps between distinct domains. Nevertheless, most DA studies concentrate on the classification settings.[8,9] They do not apply to regression molecular representation learning problems, including molecular property prediction,[10,11] 3D structure prediction,[12] molecular generation,[13] and binding affinity prediction.[14] Recently, optimal transport (OT) has been proven to be a promising tool to perform DA tasks[15] including heterogeneous DA[16] and multi-source DA.[17] However, existing OT methods to address DA of molecular representation learning are still constrained by two bottlenecks. First, the exploitation of label information in current OT methodologies is mainly designed for class labels. Second, OT for DA is based on the mini-batch training manner and the Euclidean metric. The sampled instances within mini-batches are unable to reflect the accurate distribution fully. Thus, the estimated transport plan is biased.[18]

To address the issues mentioned above, we propose an OT method called MROT, which is so designed for molecular regression DA problems (see Figure 2). In this work, we design different metrics to measure their distance across domains for the molecular representation learning of unsupervised DA (UDA) and semi-supervised DA (semi-DA). To fully utilize the regression label information of the source domain, we impose a posterior variance regularizer on the transmission plan. Apart from that, we address the problem of the large distribution by training in mini-batches and combining OT with a dynamic hierarchical triplet loss, which helps attain a more distinguishable feature space and avoids ambiguous decision boundaries. Remarkably, this loss is dynamically calculated to overcome OT's mini-batch training flaw. It explores the data distribution obtained in the previous iteration to guide the differentiation of samples in the current stage so that OT can jump out of biased local data distributions and align the domains from a global perspective. We investigate the effectiveness of our model on two sorts of tasks, namely molecular property prediction and materials adsorption selection. A broad range of compelling experiments demonstrate that MROT surpasses previous methods and can accompany DL models to achieve better generalization ability.

### DA work

DA has emerged as a new learning mechanism to address the lack of labeled data. Its purpose is to map the data of different distributions of source and target domains into the same feature space to reduce cross-domain differences. Early methods[19,20] utilize moment matching to align feature distributions, while succeeding approaches such as domain adversarial neural network (DANN),[21] conditional domain adversarial networks (CDAN),[22] and decision-boundary iterative refinement training with a teacher (DIRT-T)[23] leverage adversarial learning to diminish the domain gap. In addition, the maximum classifier discrepancy (MCD)[24] adopts prediction diversity between multiple learnable classifiers to achieve local or category-level feature alignment between source and target domains. Invariant risk minimization (IRM)[25] seeks to find optimal predictors across different such scenarios or environments simultaneously. But most prior methods are unsuitable for regression tasks and fail to prove their effectiveness in solving chemical problems.

### OT in DA

OT theory is a robust and principled approach to analyzing distances between probability distributions, with deep roots in statistics, computer science, and applied mathematics. Simultaneously, OT has also been a burgeoning tool to perform DA tasks[15,26] including heterogeneous DA[16] and multi-source DA.[17] It learns the transformation across domains under a minimal cost with theoretical guarantees.[27,28] For example, OTDA[29] seeks to find simultaneously optimal predictors across different scenarios or environments. But most prior methods are unsuitable for regression tasks and fail to prove their effectiveness in solving chemical problems. Joint distribution optimal transportation (JDOT)[30] strives to minimize the OT loss between the joint source distribution and an estimated target distribution for UDA but neglects the label information in the source domain to constrain the OT plan. While a few studies had been introduced to either learn a better metric[31,32] or reduce the bias brought by mini-batches,[18,33] none of them succeed in realizing those two objectives simultaneously.

### DA in chemical applications

DA in chemistry has raised a growing interest in the past few years. For example, splitting by scaffold or protein family may lead to a significant drop in performance.[34–37] Jin et al.[38] improve the IRM[25] with predictive regret to generalize to new scaffolds or protein families, while other methods[39] merge multiple materials datasets by introducing an additional state variable to indicate the fidelity of each dataset. Though this framework is applicable across ordered and disordered materials, it requires full access to labeled data of multiple domains. More crucially, preceding studies all leave semi-DA out of consideration, ignoring that semi-DA creates unique challenges concerning the exploitation of label information in the target domain.[40] As mentioned before, OT has shown great power in solving DA tasks. Still, most prior works apply it to learn the transformation between vision domains for classification tasks[41,42] but not chemical domains for regression tasks.

### Preliminaries
### DA for molecular representation learning

The success of supervised learning hinges on the critical hypothesis that test data ought to share the same distribution with the training data. Nonetheless, in most real-world applications, data are dynamic, meaning a distribution shift often exists between the training and test domains. Consequently, it has been universally acknowledged that deep neural networks trained on one domain can be poor at generalizing to another due to the distributional discrepancy.

DA aims to approach this problem and has been intensively researched since its first emergence.[43] Existing approaches mainly focus on classification tasks, where a classifier learns a mapping from a learned domain-invariant latent space to a fixed label space using source data. Subsequently, the classifier depends only on standard features across domains and can be applied to the target domain. For these classification problems, people regard data with class labels unseen in the source training set $y_i \notin \mathcal{Y}^s \subseteq \mathbb{Z}^+$ as the domain with a semantic shift. These categorical labels explicitly indicate their corresponding classes. In contrast, data represented in different forms are regarded as DA with a non-semantic shift. For instance, pictures can be shown in cartoons, sketches, live-action, etc.[8]

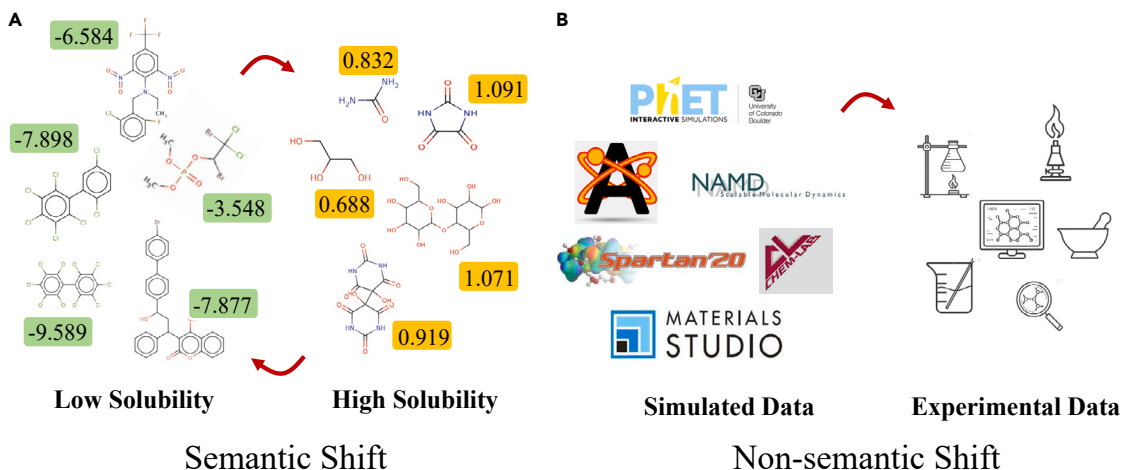

**Figure 1. Examples of the two distribution shifts in chemistry**
(A) Molecules with different ranges of properties are viewed as semantic shifts. The numbers in green and orange represent the magnitude of the compounds' solubility.
(B) Simulated and experimental data come from two domains with the non-semantic shift.

However, it is not straightforward to conceptualize the shift types for regression molecular representation learning tasks in the chemical field. To fill in this gap, we identify molecules with properties that are not in the range of the training set $y_i \notin \mathcal{Y}^s \subseteq \mathbb{R}$ as semantic shifts (see Figure 1). To be specific, the space of the property values in two different domains, $s$ and $t$, with semantic shift ought to be diverse as $\mathcal{Y}^s \neq \mathcal{Y}^t$. For instance, an adaptation from drugs with a low topological polar surface area to a high topological polar surface area is a semantic shift.[44] Besides, a molecule is identified as a non-semantic shift if drawn from an unseen category. For example, macromolecules such as polyphenols and nucleic acids can be treated as a non-semantic shift to small molecules like heteronuclear diatomic molecules. Four basic types of crystals containing covalent, ionic, metallic, and molecular crystals can also be considered a non-semantic shift from each other. Moreover, a non-semantic shift exists between the simulated and experimental data.

*Problem statement*
Throughout the article, we consider the chemical regression tasks $g : \mathcal{X} \to \mathcal{Y} \subseteq \mathbb{R}$, where $\mathbf{x} \in \mathcal{X}$ can be any small molecules, crystals, or proteins and $\mathbf{y} \in \mathcal{Y}$ represents a physical or chemical property. Typically, one assumes the existence of two distinct joint probability distributions $\mathbb{P}_s(\mathbf{x}^s, \mathbf{y}^s)$ and $\mathbb{P}_t(\mathbf{x}^t, \mathbf{y}^t)$ defined over $\mathcal{X} \times \mathcal{Y}$ and related to the source and target domains, respectively. We denote them as $\mathcal{P}^s$ and $\mathcal{P}^t$ for sake of simplicity.

Let $\mathcal{E}$ be the domain set, and $\mathcal{E}^t$ and $\mathcal{E}^s$ represent the source and target domain sets, respectively. The DA strategies can be divided into two families as semi-DA[45] and UDA,[46,47] which depend on the presence of few labels in the target domain set $\mathcal{E}^t$. $(\mathbf{x}_i^e, y_i^e)$ and $\mathcal{D}^e$ denote an input-label pair and a dataset drawn from the data distribution of some domain $e \in \mathcal{E}^s \cup \mathcal{E}^t$, respectively. The goal of DA is to seek a regressor $g_*$ that minimizes the worst-domain loss on $\mathcal{E}^t$:

$$g_* = \operatorname*{argmin}_{g \in \mathcal{G}} \mathcal{L}^{\mathcal{E}}(\mathcal{E}^t, g), \mathcal{L}^{\mathcal{E}}(\mathcal{E}, g) \triangleq \max_{e \in \mathcal{E}} \mathbb{E}\left[\ell(g(\mathbf{x}_i^e)), y_i^e)\right],$$

(Equation 1)

where $\mathcal{G} : \mathcal{X} \to \mathbb{R}$ is the hypothesis space and $\ell$ is the loss function. Furthermore, the problem is simplified in our setting, where we only consider a single source domain, $s$, and a single target domain, $t$, instead of two domain sets. Practically, we aim to minimize the error of a dataset in the target domain $\mathcal{D}^t$ as $\mathcal{L}^{\mathcal{D}}(\mathcal{D}^t, g)$, where $\mathcal{L}^{\mathcal{D}}$ corresponds to the total loss of $g$ on a given dataset $\mathcal{D}$. Similar to previous studies,[25,38,48–50] $g$ is assumed to be decomposed into $h \circ f$, where $f : \mathcal{X} \to \mathbb{R}^d$ is the feature extractor that maps the input into the feature space $\mathcal{H}$ and $h : \mathbb{R}^d \to \mathbb{R}$ is the predictor.

The supplemental experimental procedures list common assumptions made by most DA methods, including property imbalance and covariate shift, and ways to measure the variation and informativeness of $f$ also are provided.

## RESULTS

### Method overview
As shown in Figure 2, the proposed MROT consists of three parts: the mini-batch OT, the feature extractor, and the predictor. Firstly, MROT relies on mini-batch computation and aligns the distribution of different source domains via OT by minimizing the pairwise distances between elements of the source and target batches. And a posterior variance regularizer is introduced in the transmission plan to exploit the regression label information of the source domain. Then, MROT adopts the metric learning objective with dynamic hierarchical three-group loss, jumps out of the local data distribution of mini-batch training during metric learning, and considers the global data distribution of multiple domains, which helps to obtain a more easily identifiable feature space. Further details on our MROT model are given in the experimental procedures.

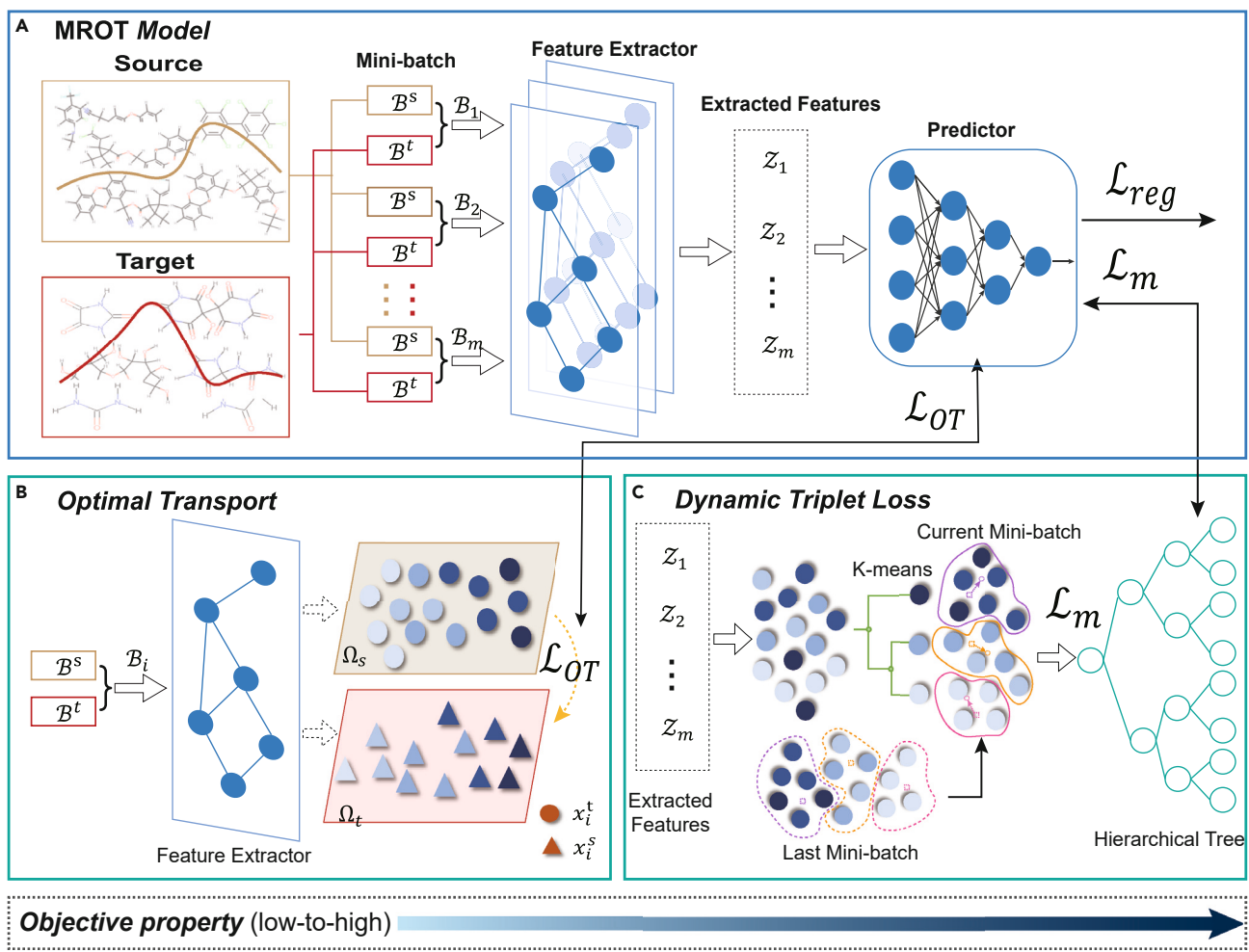

**Figure 2. The architecture of MROT**
(A) The general workflow of the MROT model.
(B) The OT process to align features from different domains.
(C) Metric learning seeks better decision boundaries with a dynamic hierarchical tree. All sample pairs with the same clusters are seen as positive, while others are regarded as negative. The darkness of blue colors reflects the magnitude of molecular properties.

### Datasets

To empirically corroborate the effectiveness of MROT, we evaluate it on real-world applications, including the molecular property prediction task (unsupervised task) and the material adsorption prediction task (semi-supervised). The description of benchmark datasets is enumerated in Table 1, including the number of tasks, the number of molecules and atom classes, the minimum and the maximum number of atoms, and their evaluation metrics.

### Molecular property prediction datasets

In the unsupervised task, to demonstrate that the MROT model can generalize beyond its training domain to make predictions, we split datasets based on the objective property instead of molecular constituents, which accords with a semantic shift. We consider a low-to-high environment rather than the high-to-low environment because molecules with high values of desired properties are traditionally what scientists or pharmacists search for in drug or materials design.[51,52] The train and test sets contain molecules with the lowest 80% and the 10% highest properties, respectively. Then, we use the remaining 10% data points for validation.

The following six regression datasets from quantum chemistry and physical chemistry were used.

- QM7[53] is a subset of GDB-13, which records the computed atomization energies of stable and synthetically accessible organic molecules.
- QM8[54] contains computer-generated quantum mechanical properties, including electronic spectra and excited-state energy of small molecules.
- QM9[10] is a comprehensive dataset that provides geometric, energetic, electronic, and thermodynamic properties for a subset of GDB-17.
- ESOL[55] is a small dataset documenting the solubility of compounds.
- FreeSolv[56] provides experimental and calculated hydration-free energy of small molecules in water. The

Table 1. Key statistics of datasets from three different chemical fields

| Category | Dataset | Tasks | Task type | Molecules | Atom class | Min. atoms | Max. atoms | Metric |
|---|---|---|---|---|---|---|---|---|
| Quantum chemistry | QM7 | 1 | regression | 7,160 | 5 | 4 | 23 | MAE |
| | QM8 | 12 | regression | 21,786 | 5 | 3 | 26 | MAE |
| | QM9 | 12 | regression | 133,885 | 5 | 3 | 28 | MAE |
| Physical chemistry | ESOL | 1 | regression | 1,128 | 9 | 1 | 55 | RMSE |
| | FreeSolv | 1 | regression | 643 | 9 | 1 | 24 | RMSE |
| | Lipophilicity | 1 | regression | 4,200 | 12 | 7 | 115 | RMSE |
| Materials science | CoRE-MOF | 6 | regression | 10,066 | 77 | 10 | 10,560 | RMSE |
| | Exp-MOF | 1 | regression | 113 | 25 | 11 | 3,594 | RMSE |

calculated values are derived from alchemical free-energy calculations using molecular dynamics simulations.
- Lipophilicity[57] (Lipo) is selected from ChEMBL, which is an important property that affects molecular membrane permeability and solubility. The data are obtained via octanol/water distribution coefficient experiments.

### Material adsorption prediction datasets

The algorithm for semi-DA is evaluated on the materials adsorption dataset. To meet the setting of semi-DA, we randomly select a ratio (25% or 50%) of Experimental metal-organic frameworks (Exp-MOF) with labeled target samples for training and use the rest as the testing set. Each method is conducted three times due to the small size of Exp-MOF, and we report the mean performance. It consists of the simulated materials data and the experimental materials data described as follows.

- Computation-ready experimental metal-organic frameworks (CoRE-MOF)[58] owns over 10,000 crystal samples and a wide variety of 79 atom classes. It is obtained from the Cambridge Structural Database[59] and a Web of Science[60] search and is derived through semi-automated reconstruction of disordered structures using a topology-based crystal generator. CoRE-MOF offers 12 chemical properties, such as carbon dioxide adsorption. Nonetheless, most of its materials are unable to be synthesized.
- Exp-MOF[61] contains samples that can be synthesized and acquired throughout rigorous and high-priced experiments. But due to the expensive costs, its data size is much smaller, with only 113 crystals. CoRE-MOF and Exp-MOF share the same target label: the adsorption capability of materials to carbon dioxide. We aspire to adapt the predictive model to forecast that property from CoRE-MOF to Exp-MOF.

### Baselines

We select five baselines for regression DA tasks. Without loss of generality, we select two sorts of popular predictive models as the backbone $f$. One is the Molformer,[62] a variant of Transformer[63] based on the self-attention mechanism. The other is the equivariant graph neural network (EGNN),[64] which extends the traditional GNN with equivariance to E(3) transformations. The predictor $h$ is a multi-layer perceptron (MLP). Other training details, including model architectures and hyperparameters, are discussed in supplemental experimental procedures S4.

- Empirical risk minimization (ERM) is trained on all available labeled data of source and, if possible, target domains.
- Domain adversarial training methods including DANN[21] and CDAN[22] seek to learn domain-invariant features.
- Meta-learning for domain generalization (MLDG)[65] is a meta-learning method that simulates domain shift by dividing training environments into meta-training and meta-testing.
- JDOT[30] is an OT framework for UDA between joint data distributions.

It is worth noting that we indicate the best method and second best method with footnotes for clear comparisons.

### Performance evaluation on two tasks

#### UDA problem

Tables 2 and 3 document the mean and standard deviation of three repetitions, where we only select five targets in QM9 that do not require thermochemical energy subtractions. MROT surpasses all baselines with significant margins. Particularly, MROT exceeds JDOT, illustrating its validity to discover better decision boundaries and overcome the drawbacks of mini-batch training. Besides, both DANN and CDAN achieve lower errors than ERM, which means that learning invariant representations can benefit UDA to some extent on the regression molecular presentation learning tasks. Remarkably, the improvement brought by MROT over other approaches in QM8 and QM9 is higher than that in ESOL, FreeSolv, and Lipo, which have less than 5,000 samples. This is because large datasets have more mini-batches, so the bias of the local data distribution in mini-batches with respect to the global data distribution is much greater. The dynamic loss empowers MROT to resist this bias. Thus, the advantage of MROT is enhanced when the data size increases.

#### Semi-DA problem

Table 4 reports the root-mean-squared error (RMSE), the Pearson correlation ($R_p$), and the Spearman correlation ($R_s$) in Table 4 with 25% and 50% labeled target data. The superiority of MROT over baseline methods is supported by its lowest mean absolute error (MAE) and highest $R_p$ and $R_s$. It is worth noting that although MLDG performs poorly in UDA, it beats adversarial methods in semi-DA, demonstrating the efficacy of meta-learning when the size of the target dataset is small.

### Ablation study and feature visualization

We examine the effects of each component in MROT. Table 5 compares the performance of ERM, OT without regularization, OT with all regularization, triplet loss only, and MROT. It demonstrates that the posterior variance regularization contributes to a substantial decrease in the adaptation error. Moreover, OT coupled with the dynamic hierarchical triplet loss produces a better performance than adopting either of them.

Table 2. Performance on small-molecule datasets with Molformer and EGNN (lower means better)

| Backbone | Method | QM7 | QM8 | ESOL | FreeSolv | Lipo |
|---|---|---|---|---|---|---|
| Molformer | ERM | 86.152 ± 1.265 | 0.013 ± 0.008 | 2.008 ± 0.014 | 0.611 ± 0.007 | 1.249 ± 0.054 |
| | DANN | 85.067 ± 2.140 | 0.008 ± 0.012 | 1.846 ± 0.023 | 0.554 ± 0.033 | 1.173 ± 0.076 |
| | CDAN | 85.203 ± 2.280 | 0.007 ± 0.014 | 1.821 ± 0.028 | 0.552 ± 0.019 | 1.165 ± 0.053 |
| | MLDG | 85.007 ± 3.390 | 0.009 ± 0.031 | 1.973 ± 0.044 | 0.585 ± 1.380 | 1.188 ± 0.081 |
| | JDOT | 84.103 ± 1.470[a] | 0.006 ± 0.005[a] | 1.805 ± 0.019[a] | 0.512 ± 0.079[a] | 1.121 ± 0.125[a] |
| | MROT | 82.599 ± 1.850[b] | 0.004 ± 0.001[b] | 1.723 ± 0.016[b] | 0.493 ± 0.046[b] | 1.104 ± 0.091[b] |
| EGNN | ERM | 58.578 ± 1.384 | 0.030 ± 0.006 | 1.633 ± 0.005 | 0.609 ± 0.038 | 1.698 ± 0.032 |
| | DANN | 55.632 ± 1.621 | 0.026 ± 0.007 | 1.429 ± 0.000 | 0.542 ± 0.042 | 1.459 ± 0.035 |
| | CDAN | 56.721 ± 1.544 | 0.027 ± 0.013 | 1.437 ± 0.008 | 0.577 ± 0.057 | 1.467 ± 0.423 |
| | MLDG | 56.196 ± 1.012 | 0.027 ± 0.024 | 1.525 ± 0.120 | 0.565 ± 0.044 | 1.522 ± 0.047 |
| | JDOT | 54.388 ± 2.890[a] | 0.022 ± 0.006[a] | 1.388 ± 0.009[a] | 0.438 ± 0.086[a] | 1.362 ± 0.184[a] |
| | MROT | 54.010 ± 2.167[b] | 0.015 ± 0.005[b] | 1.346 ± 0.015[b] | 0.421 ± 0.082[b] | 1.357 ± 0.187[b] |

[a]Second best method.
[b]Best method.

We also provide the comparison results between two different metric designs for semi-DA in Table 6. The experiment results firmly support our statement that the Jensen-Shannon (JS) distance metric outweighs the additive distance metric with generally higher $R_p$ and $R_s$ and a lower RMSE. Besides, it is also discovered that a small $\kappa$ ($\kappa = 0.2$) benefits semi-DA the most, while an extremely large $\kappa$ ($\kappa = 100$) does great harm to the performance.

We envision feature distributions of ERM and MROT in QM8 by t-distributed stochastic neighbor embedding (t-SNE) projection[66] in Figure 3. On the one hand, our approach realizes a lower MAE in the target domain, indicating its better capability of domain alignment. On the other hand, MROT succeeds at separating molecules of out-of-distribution (high) properties from molecules of in-distribution (low) properties. Thus, it can be widely applied in medicine to seek drug-like molecules with desired outstanding properties, which may never be seen in the source domain.

## DISCUSSION

Molecular representation learning is significant for drug discovery and materials synthesis fields. However, existing machine learning-based methods commonly hypothesize that the model development or experimental evaluation is mainly based on independently and identically distributed data across training and testing. In real scenarios, the available experimental molecule data are somewhat limited, while the candidate molecules are often diverse, coming from unknown environments. The adaptation across different domains guarantees the robustness of molecular representation learning models and can significantly benefit the discovery of new drugs and materials. In this work, an OT model, called MORT, is built for chemical regression DA problems with novel metrics and a posterior variance regularizer. MROT performs mini-batch operations on datasets in the calculation process to overcome the computational load brought by large data distribution, jumps out of the local data distribution of small batch training during metric learning, and considers the global data distribution of multiple domains. A dynamic hierarchical triplet loss is introduced to overcome the shortage of conventional metrics and mitigate the bias by computing mini-batches to help achieve more distinguishable decision boundaries and catch the global data distributions. We convincingly show that it can reach state-of-the-art performance on challenging unsupervised and semi-supervised tasks.

Table 3. Comparison of MAE on QM9 with Molformer and EGNN

| Backbone | Target unit | $\epsilon_{HOMO}$, eV | $\epsilon_{LUMO}$, eV | $\Delta\epsilon$, eV | $\mu$, D | $\alpha$, bohr$^3$ |
|---|---|---|---|---|---|---|
| Molformer | ERM | 0.502 ± 0.011 | 0.380 ± 0.023 | 0.501 ± 0.010 | 1.029 ± 0.019 | 3.147 ± 0.034 |
| | DANN | 0.440 ± 0.024 | 0.364 ± 0.037 | 0.443 ± 0.025 | 0.992 ± 0.037 | 2.532 ± 0.031 |
| | CDAN | 0.451 ± 0.028 | 0.361 ± 0.036 | 0.451 ± 0.033 | 0.994 ± 0.022 | 2.681 ± 0.029 |
| | MLDG | 0.465 ± 0.044 | 0.352 ± 0.049 | 0.485 ± 0.058 | 1.015 ± 0.062 | 2.695 ± 0.462 |
| | JDOT | 0.439 ± 0.018[a] | 0.311 ± 0.033[a] | 0.421 ± 0.043[a] | 0.956 ± 0.045[a] | 2.413 ± .0.036[a] |
| | MROT | 0.405 ± 0.022[b] | 0.307 ± 0.036[b] | 0.398 ± 0.044[b] | 0.923 ± 0.059[b] | 2.306 ± 0.025[b] |
| EGNN | ERM | 0.600 ± 0.015 | 0.449 ± 0.022 | 1.231 ± 0.018 | 1.876 ± 0.026 | 3.190 ± 0.033 |
| | DANN | 0.521 ± 0.037 | 0.385 ± 0.038 | 1.033 ± 0.035 | 1.562 ± 0.038 | 2.588 ± 0.041 |
| | CDAN | 0.563 ± 0.042 | 0.406 ± 0.033 | 1.026 ± 0.033 | 1.630 ± 0.033 | 2.671 ± 0.040 |
| | MLDG | 0.550 ± 0.044 | 0.417 ± 0.040 | 1.105 ± 0.047 | 1.663 ± 0.058 | 2.653 ± 0.168 |
| | JDOT | 0.498 ± 0.035[a] | 0.362 ± 0.036[a] | 0.988 ± 0.029[a] | 1.473 ± 0.040[a] | 2.480 ± 0.030[a] |
| | MROT | 0.472 ± 0.038[b] | 0.344 ± 0.033[b] | 0.927 ± 0.031[b] | 1.465 ± 0.039[b] | 2.395 ± 0.029[b] |

[a]Second best method.
[b]Best method.

Table 4. Comparison of RMSE (lower means better) and $R_p$ and $R_s$ (higher means better) on Exp-MOF with 25% and 50% labeled target data

| Backbone | Method | 25% labeled target | | | 50% labeled target | | |
|---|---|---|---|---|---|---|---|
| | | RMSE | $R_p$ | $R_s$ | RMSE | $R_p$ | $R_s$ |
| Molformer | ERM | 33.283 | 0.333 | 0.447 | 32.611 | 0.365 | 0.453 |
| | DANN | 33.266 | 0.357 | 0.436 | 30.773 | 0.380 | 0.541 |
| | CDAN | 32.239 | 0.366 | 0.450 | 30.502 | 0.383 | 0.543 |
| | MLDG | 31.403[a] | 0.330[a] | 0.430[a] | 19.666[a] | 0.431[a] | 0.541[a] |
| | MROT | 28.354[b] | 0.428[b] | 0.467[b] | 27.590[b] | 0.471[b] | 0.585[b] |
| EGNN | ERM | 28.363 | 0.262 | 0.302 | 29.045 | 0.495 | 0.353 |
| | DANN | 27.914 | 0.287 | 0.332 | 28.385 | 0.485 | 0.356 |
| | CDAN | 26.638 | 0.471 | 0.410 | 28.071 | 0.498 | 0.364 |
| | MLDG | 26.303[a] | 0.492[a] | 0.427[a] | 27.033[a] | 0.523[a] | 0.488[a] |
| | MROT | 25.607[b] | 0.513[b] | 0.461[b] | 25.030[b] | 0.545[b] | 0.511[b] |

[a]Second best method.
[b]Best method.

Notably, we do not examine the predictive performance of our model on some classic prediction tasks, such as the lipophilicity prediction problem. This is because our MORT is mainly designed for regression tasks. Our motivation is founded on that most existing mechanisms to tackle DA merely consider the classification setting, where the label space is discrete. Nevertheless, many real-world chemical applications are regression tasks, such as molecular property prediction,[62] binding affinity prediction,[14,67] molecular dynamics simulations,[68] and protein-protein docking.[69] Our work targets this category of problems and is mainly distinct from previous methods in its efficient utilization of continuous label space.

Moreover, we do not choose to split the datasets by time and argue that the time split is not a proper split standard. Indeed, some competitions, such as Critical Assessment of Structure Prediction (CASP),[70] adopt the time split to separate proteins, but this is because competitions usually take place every year or every 2 years. Researchers have to follow this paradigm to make a fair comparison with prior competition winners. But for proteins, a more widely accepted split option is to separate the proteins by sequence identity. People usually forbid the case in which the train and test sets have proteins with sequence identity higher than a pre-defined ratio (e.g., 50%). It can also be discovered that molecular datasets seldom provide time information, which shows that time split may not be necessary for data scientists. However, there are more challenging split methodologies and more complex tasks. Our settings, which adapt models from distributions of low chemical properties to high chemical properties and from simulated data to experimental data, are only an example to showcase the superiority of our method. With the light we shed, future studies will notice this essential issue and apply our technique to more professional and specific tasks in biology, chemistry, and medicine.

In addition, MROT can also be applied to other regression tasks beyond chemical domains, such as the visual domain, the language domain, and the time-series domains. It is left as future work to extend our OT algorithm to datasets in those attractive domains, where data distributions can be very different between the train and test sets.

### EXPERIMENTAL PROCEDURES

#### Resource availability

*Lead contact*
Further information and requests for resources should be directed to and will be fulfilled by the lead contact, Stan Z. Li (stan.zq.li@westlake.edu.cn).

*Materials availability*
There are no physical materials associated with this study.

Table 5. Ablation study on the components of MROT on QM7, QM8, ESOL, FreeSolv, and Lipo

| | OT | VR | TL | QM7 | QM8 | ESOL | FreeSolv | Lipo |
|---|---|---|---|---|---|---|---|---|
| 1 | – | – | – | 86.152 | 0.013 | 2.186 | 0.614 | 1.261 |
| 2 | ✓ | – | – | 84.526 | 0.009 | 1.837 | 0.579 | 1.184 |
| 3 | ✓ | ✓ | – | 83.695[a] | 0.007[a] | 1.821[a] | 0.524[a] | 1.135[a] |
| 4 | – | – | ✓ | 84.677 | 0.008 | 1.832 | 0.547 | 1.146 |
| 5 | ✓ | ✓ | ✓ | 82.599[b] | 0.004[b] | 1.723[b] | 0.493[b] | 1.104[b] |

VR, variance reduction regularization; TL, dynamic triplet loss.
[a]Second best method.
[b]Best method.

Table 6. Ablation study on the distance metric on Exp-MOF

| Method | 25% labeled target | | | 50% labeled target | | |
|---|---|---|---|---|---|---|
| | RMSE | $R_p$ | $R_s$ | RMSE | $R_p$ | $R_s$ |
| Additive distance | 29.581[a] | 0.373[a] | 0.444[a] | 28.509[a] | 0.427[a] | 0.578[a] |
| JS distance ($\kappa = 0.2$) | 28.354[b] | 0.428[b] | 0.467[b] | 27.590[b] | 0.471[b] | 0.585[b] |
| JS distance ($\kappa = 1$) | 30.410 | 0.343 | 0.409 | 31.313 | 0.358 | 0.516 |
| JS distance ($\kappa = 100$) | 39.453 | 0.117 | 0.211 | 38.769 | 0.128 | 0.208 |

[a]Second best method.
[b]Best method.

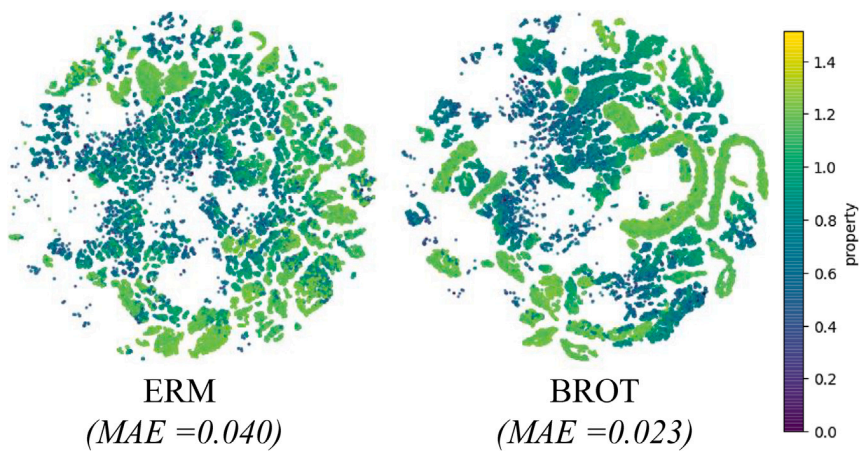

**Figure 3. The t-SNE visualization of the UDA task in QM8**

The brightness of green denotes the magnitude of molecular properties.

### Data and code availability

The code and data are available at GitHub (https://github.com/smiles724/BROT). Original data have been deposited to Zenodo.[61,71]

### Method

#### Unsupervised and semi-DA

In UDA, the ground truth of property distributions in the target domain $\mathbb{P}_t(\mathbf{y}^t)$ is inaccessible, and labels are only in the source domain. It is usually correlated with a semantic shift. The priority in UDA is the metric to measure the distance between $\mathbf{x}^s$ and $\mathbf{x}^t$ on the ground space $\mathcal{Z} = \mathcal{X}$. In most applications, the metric in the feature space, $d_{\mathcal{H}}$, is readily available, and the Euclidean distance is a popular option.[29] Consequently, the distance between two molecules $d_{\mathcal{Z}}^f : \mathcal{X} \times \mathcal{X} \to \mathbb{R}^+$ is defined as

$$d_{\mathcal{Z}}^f\left(\mathbf{x}_i^s, \mathbf{x}_j^t\right) = d_{\mathcal{H}}\left(f\left(\mathbf{x}_i^s\right), f\left(\mathbf{x}_j^t\right)\right). \quad \text{(Equation 2)}$$

Semi-DA is a more realistic setting, where learners can access a small amount of labeled data but no unlabeled data from the target domain. It usually occurs with non-semantic shift of the data distribution. To capture the mutual relationship between $\mathcal{H}$ and $\mathcal{Y}$ and avoid parameter sensitivity, we define the following metric to measure the distance between $(\mathbf{x}^s, \mathbf{y}^s)$ and $(\mathbf{x}^t, \mathbf{y}^t)$ on the ground space $\mathcal{Z} = \mathcal{X} \times \mathcal{Y}$. Concisely, we take the form of a JS divergence[72] to constrain the discrepancy between each feature-label pair as

$$d_{\mathcal{Z}}^f\left(\left(\mathbf{x}_i^s, y_i^s\right), \left(\mathbf{x}_j^t, y_j^t\right)\right) = d_{\mathcal{H}}'^P + \epsilon d_{\mathcal{Y}}'^P + \kappa \left( \left| d_{\mathcal{Y}}'^P \log\left(\frac{d_{\mathcal{H}}'^P}{d_{\mathcal{Y}}'^P + \zeta}\right) \right| + \left| d_{\mathcal{H}}'^P \log\left(\frac{d_{\mathcal{Y}}'^P}{d_{\mathcal{H}}'^P + \zeta}\right) \right| \right), \quad \text{(Equation 3)}$$

where $\epsilon$ is a hyperparameter to balance distances in two spaces $\mathcal{H}$ and $\mathcal{Y}$. $d_{\mathcal{Y}}$ is the metric in $\mathcal{Y}$. $|.|$ ensures non-negativeness. $\kappa$ is a hyperparameter. $d_{\mathcal{H}}'$ is the normalized value of $d_{\mathcal{H}}$ as $d_{\mathcal{H}}'(f(\mathbf{x}_i^s), f(\mathbf{x}_j^t)) = \frac{d_{\mathcal{H}}(f(\mathbf{x}_i^s), f(\mathbf{x}_j^t))}{d_{\mathcal{H}\ max}}$, where $d_{\mathcal{H}\ max}$ is the maximum distance of all source-target pairs in the feature space, and $d_{\mathcal{Y}}'$ is adjusted in the same way. $d_{\mathcal{H}}'$ and $d_{\mathcal{Y}}'$ are therefore between 0 and 1. $\zeta > 0$ is added to prevent the zero division error.

Particularly, Equation 3 depends on two components: the addition of $d_{\mathcal{Y}}'$ and $d_{\mathcal{H}}'$ accompanied by the JS term. The former requires the source and target samples to carry similar properties and features, while the latter imposes a strong penalty over the disagreement of $d_{\mathcal{H}}'$ and $d_{\mathcal{Y}}'$ (see Figure 4). Thus, with this JS distance metric, $d_{\mathcal{Z}}^f$ considers the magnitude of properties and features and the joint connection between two metrics synchronically.

#### Mini-batch OT with regularization

Equipped with pre-defined cost functions in $\mathcal{Z}$, our goal is to minimize a geometric notion of distance between $\mathcal{P}^s$ and $\mathcal{P}^t$. As the full OT problem is untractable for large distributions, we rely on mini-batch computation, which has recently been accommodated well with a stochastic optimization over $f$.[73] We assume that a training batch $\mathcal{B} = \mathcal{B}^s \cup \mathcal{B}^t$ contains a source batch $\mathcal{B}^s = \{(\mathbf{x}_i^s, y_i^s)\}_{i=1}^b$ and a target batch $\mathcal{B}^t = \{(\mathbf{x}_i^t, y_i^t)\}_{i=1}^b$. Explicitly, for UDA, $\mathcal{B}^t$ comes from all unlabeled data attainable in the target domain, while semi-DA is drawn only from labeled data. Here, $b$ is the mini-batch size. More formally, our objective function is

$$d_{OT}^f(\mathcal{D}^s, \mathcal{D}^t) = \mathbb{E}\left[\min_{\mathbf{T} \in \Pi(\mathcal{B}^s, \mathcal{B}^t)} \langle \mathbf{T}, \mathbf{D}_{\mathcal{Z}}^f \rangle\right], \quad \text{(Equation 4)}$$

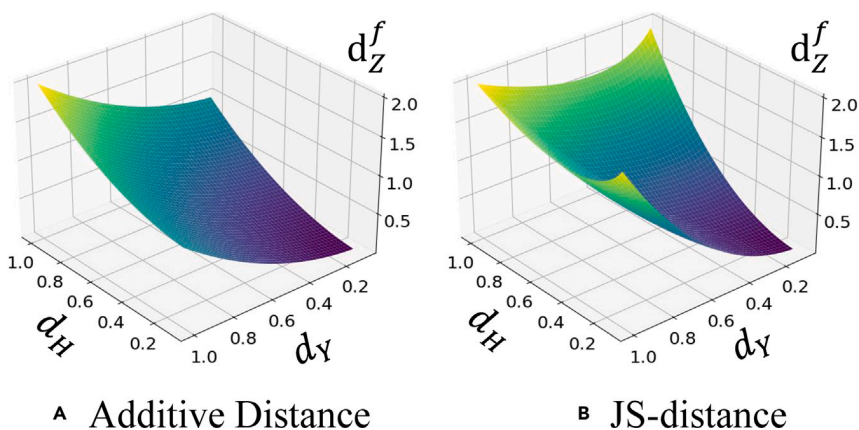

**Figure 4. The cost functions of different metrics in Semi-DA**

(A) Using the additive distance metric.
(B) Leveraging a JS-distance metric.
The x and y axes correspond to the distances in $\mathcal{H}$ and $\mathcal{Y}$.

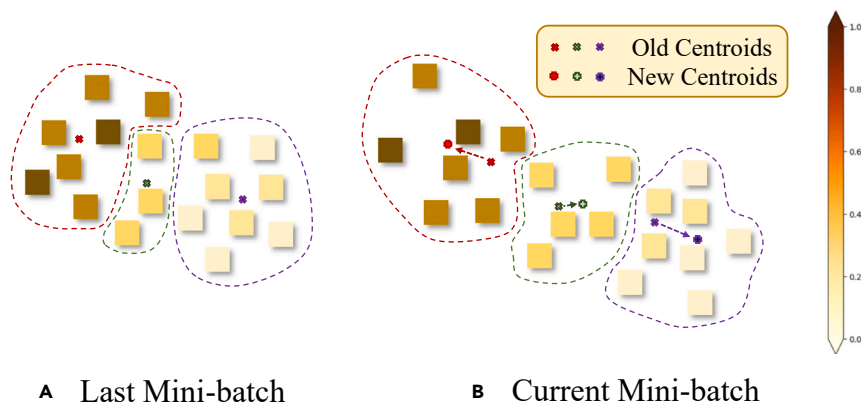

Figure 5. The dynamic cluster centroids in our triplet loss
The centroids in the last mini-batch are used to help cluster molecules in the current mini-batch and then are updated afterward. If the color is darker, the molecular property is higher.

where $D_{\mathcal{Z}}^f$ is the matrix of all pairwise distances between elements of the source and target batches, and $\mathbb{E}$ takes the expectation concerning the randomly sampled mini-batches drawn from both domains. This optimization is conducted over $\Pi(.,.)$, which is the set of all matrices with prescribed uniform marginals defined as

$$\Pi(\mathcal{B}^s, \mathcal{B}^t) = \{T \in \mathbb{R}_+^{b \times b} | T \mathbf{1}_b = \mathbf{u}_b, T^\top \mathbf{1}_b = \mathbf{u}_b\}, \quad \text{(Equation 5)}$$

where $\mathbf{u}_b$ is the uniform distribution of size $b$, and $\mathbf{1}_b \in \mathbb{R}^b$ is a vector of ones.

Besides, to enable OT to perform better in regression tasks, we overcome the problems of traditional OT with exponential dependence on data dimension and not exploiting all DA specificity by imposing a new regularization. We start by defining a posterior probability distribution of the property for the $j^{th}$ target sample and noting $Y_j$ as the corresponding random variable. $Y_j$ takes $\mathbf{y}^s$ as possible values with probabilities given by the $j^{th}$ column of the coupling matrix $T$ (noted $\mathbf{t}_j$), divided by the probability of having this sample. As batches are uniform distributions of samples, this probability vector is simply $b\mathbf{t}_j$. Intuitively, we seek to minimize the variance of the transported properties for a given sample. Therefore, for a given target sample $j$, the regularizer $\Omega_p^{y^s}$, being a function of $\mathbf{t}_j$, is defined as

$$\Omega_p^{y^s}(\mathbf{t}_j) = \mathrm{var}_{\mathbf{t}_j}(Y_j) = b \sum_{i=1}^{b} T_{ij} \left(y_i^s - \sum_{l=1}^{b} bT_{lj} \cdot y_l^s\right)^2, \quad \text{(Equation 6)}$$

where var stands for the variance. A small $\Omega_p^{y^s}$ guarantees that the $j^{th}$ target sample receives masses only from source samples with close properties and therefore induces a desired representation for them. We finally note that $\Omega_p^{y^s}(T) = \sum_j \Omega_p^{y^s}(\mathbf{t}_j)$, the total regularization over every sample of the target.

The final formulation of our problem, combining entropy and posterior variance regularization, is

$$d_{OT}^f(\mathcal{D}^s, \mathcal{D}^t) = \mathbb{E}\left[\min_{T \in \Pi(\mathcal{B}^s, \mathcal{B}^t)} \langle T, D_{\mathcal{Z}}^f \rangle + \lambda_1 \Omega_e(T) + \lambda_2 \Omega_p^{y^s}(T)\right], \quad \text{(Equation 7)}$$

where $\Omega_e(T) = \sum_{ij} T_{ij} \log T_{ij}$ is the entropy constraint. $\lambda_1 > 0$ and $\lambda_2 > 0$ are hyperparameters controlling the importance of different regularization terms. As $\Omega_p^{y^s}(T)$ is concave in $T$ (see supplemental experimental procedures S3), the corresponding optimization problem is a difference of convex program (DC). Given $f$, this problem can be solved thanks to a generalized conditional gradient approach,[29,74] and we use the corresponding Python Optimal Transport (POT) implementation.[75] More details on the optimization procedure are given in supplemental experimental procedures S2.

Used in the final loss of our problem, we need to compute $\min_f d_{OT}^f(\mathcal{D}^s, \mathcal{D}^t)$, which consists of two nested optimization problems. To compute the gradient with regards to $f$, we use the Envelope theorem[76]: since $f$ is only involved in the cost matrix $D_{\mathcal{Z}}^f$, we first compute for given mini-batches an optimal coupling $T^*$ (by solving the problem) and consider it fixed to backpropagate through the loss $\langle T^*, D_{\mathcal{Z}}^f \rangle$. This strategy is a fairly common practice when using OT in the context of DL.[15,33]

*Dynamic triplet loss for DA*
Existing OT distances including Euclidean and Wasserstein used in the cost matrix may be a suboptimal metric,[32] leading to ambiguous decision boundaries.[77] To overcome that issue, we employ metric learning to help separate the instances, promote unequivocal prediction boundaries for superior adaptation, and design a dynamic triplet loss based on a mini-batch training manner.

The goal of metric learning is to learn a distance function under semantic constraints, bringing samples of the same class closer while pushing away data of different labels.[78] However, conventional metric learning approaches require domain knowledge to classify molecules subtly. As a remedy, our dynamic triplet loss utilizes a K-means[79] algorithm to construct a hierarchical class-level tree based on molecular representations, which can naturally capture the intrinsic data distribution across domains.[80] In addition, although data structures in the feature space constantly change during the training process, the relative positions of data points are roughly preserved.[81] This phenomenon enables us to use the local data distribution gained in previous iterations to help cluster molecules in the current iteration. Specifically, the dynamic triplet loss progressively adjusts the cluster centroids during each iteration so that the information regarding the global data distribution is revealed (see Figure 5). Consequently, the learned molecular representations can jump out of local data distributions within mini-batches and capture a global data distribution of multiple domains.

*Sample clustering.* To begin with, at the initial step ($t = 0$), we partition all $2b$ observations within each mini-batch into $K$ fine-grained clusters through K-means and attain $K$ corresponding cluster centers $M^{(0)} = \{m^{(0)}\}_{i=1}^{K}$. Then, we calculate the distance matrix $D_\mathcal{C} \in \mathbb{R}^{K \times K}$ of those $K$ clusters, where the distance between the $p^{th}$ and $q^{th}$ clusters is defined as

$$d_\mathcal{C}(p, q) = \frac{1}{n_p n_q} \sum_{i \in p, j \in q} d_{\mathcal{Z}}^f((\mathbf{x}_i, y_i), (\mathbf{x}_j, y_j)), \quad \text{(Equation 8)}$$

where $n_p$ and $n_q$ are the numbers of samples belonging to the $p^{th}$ and $q^{th}$ clusters. As for UDA, $d_\mathcal{C}$ is calculated based on $d_{\mathcal{Z}}^f(\mathbf{x}_i, \mathbf{x}_j)$, and we deemphasize this difference in the rest of this section. After that, a hierarchical tree $U^{(0)}$ is created by recursively merging the leave nodes at different levels according to the distance matrix $D_\mathcal{C}$.[82]

*Triplet loss.* After we group data points in a mini-batch into different clusters, triplets are then constructed as $\psi = (\mathbf{x}_{anc}, \mathbf{x}_{pos}, \mathbf{x}_{neg})$, which contains an anchor sample $x_{anc}$, a positive sample $x_{pos}$, and a negative sample $x_{neg}$. Finally, the triplet loss can be formulated as

$$\mathcal{L}_m = \frac{1}{|\Psi^\mathcal{B}|} \sum_{\psi \in \Psi} \left[d_{\mathcal{Z}}^f((\mathbf{x}_{anc}, y_{anc}), (\mathbf{x}_{pos}, y_{pos})) - d_{\mathcal{Z}}^f((\mathbf{x}_{anc}, y_{anc}), (\mathbf{x}_{neg}, y_{neg})) + \mu\right]_+, \quad \text{(Equation 9)}$$

where $[.]_+ = \max(., 0)$ is the ramp function and $\Psi^\mathcal{B}$ and $|\Psi^\mathcal{B}|$ are the set and number of all triplets in the mini-batch $\mathcal{B}$, respectively. $d_{\mathcal{Z}}^f((\mathbf{x}_{anc}, y_{anc}), (\mathbf{x}_{pos}, y_{pos}))$ and $d_{\mathcal{Z}}^f((\mathbf{x}_{anc}, y_{anc}), (\mathbf{x}_{neg}, y_{neg}))$ separately calculate the distance of positive pairs and negative pairs. $\mu m$ is a hierarchical violate margin,[81] different from the constant margin of the conventional triplet loss. It is computed by the relationship between the centroid $m_{anc}^{(0)}$ that the anchor belongs to and the centroid $m_{neg}^{(0)}$ that is related to the negative sample, which takes the following form as

**Algorithm 1. The workflow of MROT**

> Require: A source dataset $D^s$ with $N_s$ samples and a target dataset $D^t$ with $N_t$ samples. A pre-defined number of all clusters $K$.
>
> Ensure: Model parameters $\theta_g$.
>
> $M^{(0)} \leftarrow$ K-means $(\mathcal{B}^s_{(0)}, \mathcal{B}^t_{(0)})$ ▷ initialize centroids,
>
> for $t = 1, \ldots, \lceil \frac{N_s}{b} \rceil - 1$ do
>
> $\quad \{S_i^{(t)}\}_{i=1}^K \leftarrow$ cluster $\mathcal{B}^s_{(t)}$ and $\mathcal{B}^t_{(t)}$,
>
> $\quad M^{(t)} \leftarrow$ update $M^{(t-1)}$,
>
> $\quad$ construct a hierarchical tree $\Upsilon^{(t)}$ by Equation 8,
>
> $\quad$ calculate the regression loss $\mathcal{L}_{reg}$,
>
> $\quad$ calculate the loss $\mathcal{L}_m$ by Equation 9,
>
> $\quad$ compute the optimal coupling $T^*$ from Equation 7,
>
> $\quad$ compute the total loss $\mathcal{L}$ by Equation 13,
>
> $\quad \theta_g \leftarrow \theta_g - \nabla_{\theta_g} \mathcal{L}$.
>
> end for
>
> return $\theta_g^*$

$$\mu = \mu_0 + d_\Psi\left(m_{anc}^{(0)}, m_{neg}^{(0)}\right), \quad \text{(Equation 10)}$$

where $\mu_0$ is a small constant that encourages molecular clusters to reside further apart than in previous iterations. $d_\Psi(p,q)$ is the threshold for merging the $p^{th}$ and $q^{th}$ clusters into a single node of the next level. It measures the minimal distance between clusters in the hierarchical tree $\Upsilon^{(0)}$. Notably, $\mathcal{L}_m$ computes all possible triplets within the batch.

*Iteration for mini-batch training*

In the following iterations ($t > 0$), the training proceeds by alternating between the assignment and update steps. In the assignment step, samples of a new mini-batch are allocated to $K$ clusters based on their distances to previous centroids $M^{(t-1)}$. The new $p^{th}$ cluster $S_p^{(t)}$ can be represented as

$$\left\{ x_i : d_Z^f\left((x_i, y_i), m_p^{(t-1)}\right) \leq d_Z^f\left((x_i, y_i), m_q^{(t-1)}\right), \forall q \right\}. \quad \text{(Equation 11)}$$

In the update step, the centroids $M^{(t)}$ are recalculated by aggregating the means of molecules within this mini-batch assigned to each cluster as

$$M^{(t)} = \left\{ m_p^{(t)} = \frac{1}{|S_p^{(t)}|} \sum_{x_i \in S_p^{(t)}} (x_i, y_i) \right\}_{p=1}^K. \quad \text{(Equation 12)}$$

At the same time, a new hierarchical tree $\Upsilon^{(t)}$ is reconstructed according to those new clusters. With $\Upsilon^{(t)}$, the triplet loss $\mathcal{L}_m$ within this mini-batch can also be computed by Equation 9. As the training steps proceed, $M^{(t)}$ are dynamically adjusted, and therefore the triplet loss $\mathcal{L}_m$ varies along with the changing structure of $\Upsilon^{(t)}$.

*Training loss and overall workflow*

The whole loss function of MROT per mini-batch consists of three parts, namely the regression task loss $\mathcal{L}_{reg}$, the OT loss $\mathcal{L}_{OT}$, and the metric learning loss $\mathcal{L}_m$, which can be written as

$$\mathcal{L} = \mathcal{L}_{reg} + \alpha \mathcal{L}_{OT} + \beta \mathcal{L}_m, \quad \text{(Equation 13)}$$

where $\alpha$ and $\beta$ are used to balance the effects of these three loss terms, and $\mathcal{L}_{OT}$ represents the transport distance $d_{OT}^f(\mathcal{B}^s, \mathcal{B}^t)$. Notably, $\mathcal{L}_{reg}$ contains the loss of labeled data from both the source and target domain for semi-DA (see supplemental experimental procedures S3 for more details). The complete training process is shown in Algorithm 1, where we omit the loss in the first mini-batch.

### SUPPLEMENTAL INFORMATION

Supplemental information can be found online at https://doi.org/10.1016/j.patter.2023.100714.

### ACKNOWLEDGMENTS

We thank Doctor Jiyu Cui at Zhejiang University for providing valuable material datasets. This work was supported by National Key R&D Program of China (no. 2022ZD0115100), National Natural Science Foundation of China Project (no. U21A20427), and Project (no. WU2022A009) from the Center of Synthetic Biology and Integrated Bioengineering of Westlake University.

### AUTHOR CONTRIBUTIONS

F.W. and N.C. led the research. F.W. contributed technical ideas. F.W. and N.C. developed the proposed method. F.W., N.C., and S.J. performed the analysis. S.Z.L. provided evaluation and suggestions. All authors contributed to the manuscript.

### DECLARATION OF INTERESTS

The authors have declared no competing interests.

Received: August 30, 2022
Revised: December 29, 2022
Accepted: March 1, 2023
Published: March 29, 2023

### REFERENCES

1. Goh, G.B., Hodas, N.O., and Vishnu, A. (2017). Deep learning for computational chemistry. J. Comput. Chem. 38, 1291–1307. https://doi.org/10.1002/jcc.24764.

2. Yang, S., Li, Z., Song, G., and Cai, L. (2021). Deep molecular representation learning via fusing physical and chemical information. Adv. Neural Inf. Process. Syst. 34, 16346–16357.

3. Fang, X., Liu, L., Lei, J., He, D., Zhang, S., Zhou, J., Wang, F., Wu, H., and Wang, H. (2022). Geometry-enhanced molecular representation learning for property prediction. Nat. Mach. Intell. 4, 127–134. https://doi.org/10.1038/s42256-021-00438-4.

4. Guo, Z., Nan, B., Tian, Y., Wiest, O., Zhang, C., and Chawla, N.V. (2022). Graph-based molecular representation learning. Preprint at arXiv. https://doi.org/10.48550/arXiv.2207.04869.

5. Townshend, R.J., Vögele, M., Suriana, P., Derry, A., Powers, A., Laloudakis, Y., Balachandar, S., Jing, B., Anderson, B., Eismann, S., et al. (2020). Atom3d: tasks on molecules in three dimensions. Preprint at arXiv. https://doi.org/10.48550/arXiv.2012.04035.

6. Yang, J., Zhou, K., Li, Y., and Liu, Z. (2021). Generalized out-of-distribution detection: a survey. Preprint at arXiv. https://doi.org/10.48550/arXiv.2110.11334.
7. Han, K., Lakshminarayanan, B., and Liu, J. (2021). Reliable graph neural networks for drug discovery under distributional shift. Preprint at arXiv. https://doi.org/10.48550/arXiv.2111.12951.
8. Hsu, Y.-C., Shen, Y., Jin, H., and Kira, Z. (2020). Generalized odin: detecting out-of-distribution image without learning from out-of-distribution data. In Proceedings of the IEEE/CVF Conference on Computer Vision and Pattern Recognition, pp. 10951–10960.
9. Zhang, C., Cai, Y., Lin, G., and Shen, C. (2020). Deepemd: few-shot image classification with differentiable earth mover's distance and structured classifiers. In Proceedings of the IEEE/CVF Conference on Computer Vision and Pattern Recognition, pp. 12203–12213.
10. Ramakrishnan, R., Dral, P.O., Rupp, M., and Von Lilienfeld, O.A. (2014). Quantum chemistry structures and properties of 134 kilo molecules. Sci. Data *1*, 140022. https://doi.org/10.1038/sdata.2014.22.
11. Wu, Z., Ramsundar, B., Feinberg, E.N., Gomes, J., Geniesse, C., Pappu, A.S., Leswing, K., and Pande, V. (2018). Moleculenet: a benchmark for molecular machine learning. Chem. Sci. *9*, 513–530. https://doi.org/10.1039/C7SC02664A.
12. Jumper, J., Evans, R., Pritzel, A., Green, T., Figurnov, M., Ronneberger, O., Tunyasuvunakool, K., Bates, R., Žídek, A., Potapenko, A., et al. (2021). Highly accurate protein structure prediction with alphafold. Nature *596*, 583–589. https://doi.org/10.1038/s41586-021-03819-2.
13. Madani, A., McCann, B., Naik, N., Keskar, N.S., Anand, N., Eguchi, R.R., Huang, P.-S., and Socher, R. (2020). Progen: language modeling for protein generation. Preprint at arXiv. https://doi.org/10.48550/arXiv.2004.03497.
14. Wang, R., Fang, X., Lu, Y., Yang, C.-Y., and Wang, S. (2005). The pdbbind database: methodologies and updates. J. Med. Chem. *48*, 4111–4119. https://doi.org/10.1021/jm048957q.
15. Damodaran, B.B., Kellenberger, B., Flamary, R., Tuia, D., and Courty, N. (2018). Deepjdot: deep joint distribution optimal transport for unsupervised domain adaptation. In Proceedings of the European Conference on Computer Vision (ECCV), pp. 447–463.
16. Yan, Y., Li, W., Wu, H., Min, H., Tan, M., and Wu, Q. (2018). Semi-supervised optimal transport for heterogeneous domain adaptation. IJCAI *7*, 2969–2975.
17. Nguyen, T., Le, T., Zhao, H., Tran, Q.H., Nguyen, T., and Phung, D. (2021). Most: multi-source domain adaptation via optimal transport for student-teacher learning. In Uncertainty in Artificial Intelligence (PMLR), pp. 225–235.
18. Li, M., Zhai, Y.-M., Luo, Y.-W., Ge, P.-F., and Ren, C.-X. (2020). Enhanced transport distance for unsupervised domain adaptation. In Proceedings of the IEEE/CVF Conference on Computer Vision and Pattern Recognition, pp. 13936–13944.
19. Long, M., Cao, Y., Wang, J., and Jordan, M. (2015). Learning transferable features with deep adaptation networks. In International Conference on Machine Learning (PMLR), pp. 97–105.
20. Sun, B., Feng, J., and Saenko, K. (2016). Return of frustratingly easy domain adaptation. In Proceedings of the AAAI Conference on Artificial Intelligence, *30*. https://doi.org/10.1609/aaai.v30i1.10306.
21. Ganin, Y., Ustinova, E., Ajakan, H., Germain, P., Larochelle, H., Laviolette, F., Marchand, M., and Lempitsky, V. (2016). Domain-adversarial training of neural networks. J. Mach. Learn. Res. *17*, 2096–2030.
22. Long, M., Cao, Z., Wang, J., and Jordan, M.I. (2018). Conditional adversarial domain adaptation. Adv. Neural Inf. Process. Syst. *31*.
23. Shu, R., Bui, H.H., Narui, H., and Ermon, S. (2018). A dirt-t approach to unsupervised domain adaptation. Preprint at arXiv. https://doi.org/10.48550/arXiv.1802.08735.
24. Saito, K., Watanabe, K., Ushiku, Y., and Harada, T. (2018). Maximum classifier discrepancy for unsupervised domain adaptation. In Proceedings of the IEEE Conference on Computer Vision and Pattern Recognition, pp. 3723–3732. https://doi.org/10.48550/arXiv.1802.08735.
25. Arjovsky, M., Bottou, L., Gulrajani, I., and Lopez-Paz, D. (2019). Invariant risk minimization. Preprint at arXiv. https://doi.org/10.48550/arXiv.1907.02893.
26. Sun, B., Feng, J., and Saenko, K. (2017). Correlation alignment for unsupervised domain adaptation. In Domain Adaptation in Computer Vision Applications (Cham: Springer), pp. 153–171. https://doi.org/10.1007/978-3-319-58347-1_8.
27. Seguy, V., Damodaran, B.B., Flamary, R., Courty, N., Rolet, A., and Blondel, M. (2017). Large-scale optimal transport and mapping estimation. Preprint at arXiv. https://doi.org/10.48550/arXiv.1711.02283.
28. Redko, I., Habrard, A., and Sebban, M. (2017). Theoretical analysis of domain adaptation with optimal transport. In Joint European Conference on Machine Learning and Knowledge Discovery in Databases (Springer), pp. 737–753. https://doi.org/10.1007/978-3-319-71246-8_45.
29. Courty, N., Flamary, R., Tuia, D., and Rakotomamonjy, A. (2017). Optimal transport for domain adaptation. IEEE Trans. Pattern Anal. Mach. Intell. *39*, 1853–1865.
30. Courty, N., Flamary, R., Habrard, A., and Rakotomamonjy, A. (2017). Joint distribution optimal transportation for domain adaptation. Adv. Neural Inf. Process. Syst. *30*.
31. Dhouib, S., Redko, I., Kerdoncuff, T., Emonet, R., and Sebban, M. (2020). A swiss army knife for minimax optimal transport. In International Conference on Machine Learning, pp. 2504–2513.
32. Kerdoncuff, T., Emonet, R., and Sebban, M. (2021). Metric learning in optimal transport for domain adaptation. In International Joint Conference on Artificial Intelligence.
33. Fatras, K., Séjourné, T., Flamary, R., and Courty, N. (2021). Unbalanced minibatch optimal transport; applications to domain adaptation. In International Conference on Machine Learning (PMLR), pp. 3186–3197.
34. Hou, J., Adhikari, B., and Cheng, J. (2018). Deepsf: deep convolutional neural network for mapping protein sequences to folds. Bioinformatics *34*, 1295–1303. https://doi.org/10.1093/bioinformatics/btx780.
35. Yang, K., Swanson, K., Jin, W., Coley, C., Eiden, P., Gao, H., Guzman-Perez, A., Hopper, T., Kelley, B., Mathea, M., et al. (2019). Analyzing learned molecular representations for property prediction. J. Chem. Inf. Model. *59*, 3370–3388. https://doi.org/10.1021/acs.jcim.9b00237.
36. Rao, R., Bhattacharya, N., Thomas, N., Duan, Y., Chen, X., Canny, J., Abbeel, P., and Song, Y.S. (2019). Evaluating protein transfer learning with tape. Adv. Neural Inf. Process. Syst. *32*, 9689–9701.
37. Feinberg, E.N., Sur, D., Wu, Z., Husic, B.E., Mai, H., Li, Y., Sun, S., Yang, J., Ramsundar, B., and Pande, V.S. (2018). Potentialnet for molecular property prediction. ACS Cent. Sci. *4*, 1520–1530. https://doi.org/10.1021/acscentsci.8b00507.
38. Jin, W., Barzilay, R., and Jaakkola, T. (2020). Enforcing predictive invariance across structured biomedical domains. Preprint at arXiv. https://doi.org/10.48550/arXiv.2006.03908.
39. Chen, C., Zuo, Y., Ye, W., Li, X., and Ong, S.P. (2021). Learning properties of ordered and disordered materials from multi-fidelity data. Nat. Comput. Sci. *1*, 46–53. https://doi.org/10.1038/s43588-020-00002-x.
40. Saito, K., Kim, D., Sclaroff, S., Darrell, T., and Saenko, K. (2019). Semi-supervised domain adaptation via minimax entropy. In Proceedings of the IEEE/CVF International Conference on Computer Vision, pp. 8050–8058.
41. Courty, N., Flamary, R., and Tuia, D. (2014). Domain adaptation with regularized optimal transport. In Joint European Conference on Machine Learning and Knowledge Discovery in Databases (Springer), pp. 274–289. https://doi.org/10.1007/978-3-662-44848-9_18.
42. Perrot, M., Courty, N., Flamary, R., and Habrard, A. (2016). Mapping estimation for discrete optimal transport. Adv. Neural Inf. Process. Syst. *29*, 4197–4205.
43. Hendrycks, D., and Gimpel, K. (2016). A baseline for detecting misclassified and out-of-distribution examples in neural networks. Preprint at arXiv. https://doi.org/10.48550/arXiv.1610.02136.
44. Ertl, P., Rohde, B., and Selzer, P. (2000). Fast calculation of molecular polar surface area as a sum of fragment-based contributions and its

44. application to the prediction of drug transport properties. J. Med. Chem. *43*, 3714–3717. 10.1021/jm000942e. https://doi.org/10.1021/jm000942e.
45. Kulis, B., Saenko, K., and Darrell, T. (2011). What you saw is not what you get: domain adaptation using asymmetric kernel transforms. In CVPR 2011 (IEEE), pp. 1785–1792. https://doi.org/10.1109/CVPR.2011.5995702.
46. Gopalan, R., Li, R., and Chellappa, R. (2011). Domain adaptation for object recognition: an unsupervised approach. In 2011 International Conference on Computer Vision (IEEE), pp. 999–1006.
47. Gong, B., Shi, Y., Sha, F., and Grauman, K. (2012). Geodesic flow kernel for unsupervised domain adaptation. In 2012 IEEE Conference on Computer Vision and Pattern Recognition (IEEE), pp. 2066–2073. https://doi.org/10.1109/CVPR.2012.6247911.
48. Creager, E., Jacobsen, J.-H., and Zemel, R. (2020). Exchanging Lessons between Algorithmic Fairness and Domain Generalization.
49. Krueger, D., Caballero, E., Jacobsen, J.-H., Zhang, A., Binas, J., Zhang, D., Le Priol, R., and Courville, A. (2021). Out-of-distribution generalization via risk extrapolation (rex). In International Conference on Machine Learning (PMLR), pp. 5815–5826.
50. Ye, H., Xie, C., Cai, T., Li, R., Li, Z., and Wang, L. (2021). Towards a theoretical framework of out-of-distribution generalization. Adv. Neural Inf. Process. Syst. *34*, 23519–23531.
51. Gómez-Bombarelli, R., Wei, J.N., Duvenaud, D., Hernández-Lobato, J.M., Sánchez-Lengeling, B., Sheberla, D., Aguilera-Iparraguirre, J., Hirzel, T.D., Adams, R.P., and Aspuru-Guzik, A. (2018). Automatic chemical design using a data-driven continuous representation of molecules. ACS Cent. Sci. *4*, 268–276. https://doi.org/10.1021/acscentsci.7b00572.
52. Sanchez-Lengeling, B., and Aspuru-Guzik, A. (2018). Inverse molecular design using machine learning: generative models for matter engineering. Science *361*, 360–365. https://doi.org/10.1126/science.aat26.
53. Blum, L.C., and Reymond, J.-L. (2009). 970 million druglike small molecules for virtual screening in the chemical universe database gdb-13. J. Am. Chem. Soc. *131*, 8732–8733. https://doi.org/10.1021/ja902302h.
54. Ramakrishnan, R., Hartmann, M., Tapavicza, E., and Von Lilienfeld, O.A. (2015). Electronic spectra from tddft and machine learning in chemical space. J. Chem. Phys. *143*, 084111. https://doi.org/10.1063/1.4928757.
55. Delaney, J.S. (2004). Esol: estimating aqueous solubility directly from molecular structure. J. Chem. Inf. Comput. Sci. *44*, 1000–1005. https://doi.org/10.1021/ci034243x.
56. Mobley, D.L., and Guthrie, J.P. (2014). Freesolv: a database of experimental and calculated hydration free energies, with input files. J. Comput. Aided Mol. Des. *28*, 711–720. https://doi.org/10.1007/s10822-014-9747-x.
57. Gaulton, A., Bellis, L.J., Bento, A.P., Chambers, J., Davies, M., Hersey, A., Light, Y., McGlinchey, S., Michalovich, D., Al-Lazikani, B., and Overington, J.P. (2012). Chembl: a large-scale bioactivity database for drug discovery. Nucleic Acids Res. *40*, 1100–1107. https://doi.org/10.1093/nar/gkr777.
58. Chung, Y.G., Haldoupis, E., Bucior, B.J., Haranczyk, M., Lee, S., Zhang, H., Vogiatzis, K.D., Milisavljevic, M., Ling, S., Camp, J.S., et al. (2019). Advances, updates, and analytics for the computation-ready, experimental metal–organic framework database: core mof 2019. J. Chem. Eng. Data *64*, 5985–5998. https://doi.org/10.1021/acs.jced.9b00835.
59. Groom, C.R., Bruno, I.J., Lightfoot, M.P., and Ward, S.C. (2016). The cambridge structural database. Acta Crystallogr. B Struct. Sci. Cryst. Eng. Mater. *72*, 171–179. https://doi.org/10.1107/S2052520616003954.
60. Analytics, C. (2017). Web of science core collection. Citation database. Web of Science.
61. Wu, F. (2023). Material Absorption Data for MROT. (Zenodo). https://doi.org/10.5281/zenodo.7601765.
62. Wu, F., Zhang, Q., Radev, D., Cui, J., Zhang, W., Xing, H., Zhang, N., and Chen, H. (2021). 3d-transformer: molecular representation with transformer in 3d space. Preprint at arXiv. https://doi.org/10.48550/arXiv.2110.01191.
63. Vaswani, A., Shazeer, N., Parmar, N., Uszkoreit, J., Jones, L., Gomez, A.N., Kaiser, Ł., and Polosukhin, I. (2017). Attention is all you need. In Advances in Neural Information Processing Systems, pp. 5998–6008.
64. Satorras, V.G., Hoogeboom, E., and Welling, M. (2021). E (n) equivariant graph neural networks. In International Conference on Machine Learning (PMLR), pp. 9323–9332.
65. Li, D., Yang, Y., Song, Y.-Z., and Hospedales, T.M. (2018). Learning to generalize: meta-learning for domain generalization. In Thirty-Second AAAI Conference on Artificial Intelligence. https://doi.org/10.1609/aaai.v32i1.11596.
66. Van der Maaten, L., and Hinton, G. (2008). Visualizing data using t-sne. J. Mach. Learn. Res. *9*.
67. Wu, F., Jin, S., Jiang, Y., Jin, X., Tang, B., Niu, Z., Liu, X., Zhang, Q., Zeng, X., and Li, S.Z. (2022). Pre-training of equivariant graph matching networks with conformation flexibility for drug binding. Adv. Sci. *9*, 2203796. https://doi.org/10.1002/advs.202203796.
68. Wu, F., Zhang, Q., Jin, X., Jiang, Y., and Li, S.Z. (2022). A score-based geometric model for molecular dynamics simulations. Preprint at arXiv. https://doi.org/10.48550/arXiv.2204.08672.
69. Ganea, O.-E., Huang, X., Bunne, C., Bian, Y., Barzilay, R., Jaakkola, T., and Krause, A. (2021). Independent se (3)-equivariant models for end-to-end rigid protein docking. Preprint at arXiv. https://doi.org/10.48550/arXiv.2111.07786.
70. Kryshtafovych, A., Schwede, T., Topf, M., Fidelis, K., and Moult, J. (2019). Critical assessment of methods of protein structure prediction (casp)—round xiii. Proteins *87*, 1011–1020. https://doi.org/10.1002/prot.25823.
71. smiles724. (2023). smiles724/MROT: v1.0.0. Version 1.0.0 (Zenodo). https://doi.org/10.5281/zenodo.7612040.
72. Lin, J. (1991). Divergence measures based on the shannon entropy. IEEE Trans. Inf. Theory *37*, 145–151. https://doi.org/10.1109/18.61115.
73. Fatras, K., Zine, Y., Majewski, S., Flamary, R., Gribonval, R., and Courty, N. (2021). Minibatch optimal transport distances; analysis and applications. Preprint at arXiv. https://doi.org/10.48550/arXiv.2101.01792.
74. Rakotomamonjy, A., Flamary, R., and Courty, N. (2015). Generalized conditional gradient: analysis of convergence and applications. Preprint at arXiv. https://doi.org/10.48550/arXiv.1510.06567.
75. Flamary, R., Courty, N., Gramfort, A., Alaya, M.Z., Boisbunon, A., Chambon, S., Chapel, L., Corenflos, A., Fatras, K., Fournier, N., et al. (2021). Pot: Python optimal transport. J. Mach. Learn. Res. *22*, 1–8.
76. Bonnans, J.F., and Shapiro, A. (1998). Optimization problems with perturbations: a guided tour. SIAM Rev. Soc. Ind. Appl. Math. *40*, 228–264. https://doi.org/10.1137/S0036144596302644.
77. Dou, Q., Coelho de Castro, D., Kamnitsas, K., and Glocker, B. (2019). Domain generalization via model-agnostic learning of semantic features. Adv. Neural Inf. Process. Syst. *32*, 6450–6461.
78. Kulis, B. (2013). Metric learning: a survey. In Foundations and Trends® in Machine Learning, *5*, pp. 287–364. https://doi.org/10.1561/2200000019.
79. Hartigan, J.A., and Wong, M.A. (1979). Algorithm as 136: a k-means clustering algorithm. Appl. Stat. *28*, 100–108. https://doi.org/10.2307/2346830.
80. Langfelder, P., Zhang, B., and Horvath, S. (2008). Defining clusters from a hierarchical cluster tree: the dynamic tree cut package for r. Bioinformatics *24*, 719–720. https://doi.org/10.1093/bioinformatics/btm563.
81. Ge, W. (2018). Deep metric learning with hierarchical triplet loss. In Proceedings of the European Conference on Computer Vision (ECCV), pp. 269–285.
82. Chen, T.-S., Tsai, T.-H., Chen, Y.-T., Lin, C.-C., Chen, R.-C., Li, S.-Y., and Chen, H.-Y. (2005). A combined k-means and hierarchical clustering method for improving the clustering efficiency of microarray. In 2005 International Symposium on Intelligent Signal Processing and Communication Systems (IEEE), pp. 405–408. https://doi.org/10.1109/ISPACS.2005.1595432.